\def\year{2019}\relax
\documentclass[letterpaper]{article} %DO NOT CHANGE THIS
\usepackage{aaai19}  %Required
\usepackage{times}  %Required
\usepackage{helvet}  %Required
\usepackage{courier}  %Required
\usepackage{url}  %Required
\usepackage{graphicx}  %Required
\newtheorem{THEOREM}{Theorem}[section]
\newenvironment{theorem}{\begin{THEOREM} \hspace{-.85em} {\bf :} }%
                        {\end{THEOREM}}
\newtheorem{LEMMA}[THEOREM]{Lemma}
\newenvironment{lemma}{\begin{LEMMA} \hspace{-.85em} {\bf :} }%
                      {\end{LEMMA}}
\newtheorem{COROLLARY}[THEOREM]{Corollary}
\newenvironment{corollary}{\begin{COROLLARY} \hspace{-.85em} {\bf :} }%
                          {\end{COROLLARY}}
\newtheorem{PROPOSITION}[THEOREM]{Proposition}
\newenvironment{proposition}{\begin{PROPOSITION} \hspace{-.85em} {\bf :} }%
                            {\end{PROPOSITION}}
\newtheorem{DEFINITION}[THEOREM]{Definition}
\newenvironment{definition}{\begin{DEFINITION} \hspace{-.85em} {\bf :} \rm}%
                            {\end{DEFINITION}}
\newtheorem{CLAIM}[THEOREM]{Claim}
\newenvironment{claim}{\begin{CLAIM} \hspace{-.85em} {\bf :} \rm}%
                            {\end{CLAIM}}
\newtheorem{EXAMPLE}[THEOREM]{Example}
\newenvironment{example}{\begin{EXAMPLE} \hspace{-.85em} {\bf :} \rm}%
                            {\end{EXAMPLE}}
\newtheorem{REMARK}[THEOREM]{Remark}
\newenvironment{remark}{\begin{REMARK} \hspace{-.85em} {\bf :} \rm}%
                            {\end{REMARK}}
\newcommand{\thm}{\begin{theorem}}
\newcommand{\lem}{\begin{lemma}}
\newcommand{\pro}{\begin{proposition}}
\newcommand{\dfn}{\begin{definition}}
\newcommand{\rem}{\begin{remark}}
\newcommand{\xam}{\begin{example}}
\newcommand{\cor}{\begin{corollary}}
\newcommand{\ethm}{\end{theorem}}
\newcommand{\elem}{\end{lemma}}
\newcommand{\epro}{\end{proposition}}
\newcommand{\edfn}{\bbox\end{definition}}
\newcommand{\erem}{\bbox\end{remark}}
\newcommand{\exam}{\bbox\end{example}}
\newcommand{\ecor}{\end{corollary}}
\newcommand{\eprf}{\bbox\vspace{0.1in}}
\newcommand{\beqn}{\begin{equation}}
\newcommand{\eeqn}{\end{equation}}
\newcommand{\bbox}{\vrule height7pt width4pt depth1pt}

\newcommand{\clm}{\begin{claim}}
\newcommand{\eclm}{\end{claim}}
\newcommand{\sat}{\models}

\newcommand{\union}{\cup}
\newcommand{\inter}{\cap}
\renewcommand{\phi}{\varphi}
\newcommand{\F}{{\cal F}}

\newcommand{\I}{{\cal I}}

\newcommand{\R}{{\cal R}}

\newcommand{\U}{{\cal U}}
\newcommand{\V}{{\cal V}}

\newcommand{\ol}{\setlength{\itemsep}{0pt}\begin{enumerate}}
\newcommand{\eol}{\end{enumerate}\setlength{\itemsep}{-\parsep}}
\newcommand{\ul}{\setlength{\itemsep}{0pt}\begin{itemize}}
\newcommand{\dl}{\setlength{\itemsep}{0pt}\begin{description}}
\newcommand{\edl}{\end{description}\setlength{\itemsep}{-\parsep}}
\newcommand{\eul}{\end{itemize}\setlength{\itemsep}{-\parsep}}

\newcommand{\commentout}[1]{}

\newcommand{\bi}{\begin{itemize}}
\newcommand{\ei}{\end{itemize}}
\newcommand{\be}{\begin{enumerate}}
\newcommand{\ee}{\end{enumerate}}

\usepackage{amssymb}
\usepackage{amsmath}
\usepackage{graphicx} % for pdf, bitmapped graphics files
\usepackage{subfigure}

\newcommand{\fullv}[1]{#1}
\newcommand{\shortv}{\commentout}

\newcommand{\Rest}{\mathit{Rst}}
\newcommand{\UH}{\mathit{UH}}
\newcommand{\LH}{\mathit{LH}}
\newcommand{\ULH}{\mathit{ULH}} 
\begin{document}
% The file aaai.sty is the style file for AAAI Press 
% proceedings, working notes, and technical reports.
%
\title{Abstracting Causal Models}
\author{Sander Beckers\\
Dept. of Philosophy and Religious Studies\\
Utrecht University\\
Utrecht, Netherlands\\
srekcebrednas@gmail.com
\And
Joseph Y. Halpern\\
Dept. of Computer Science\\
Cornell University\\
Ithaca, NY 14853\\
halpern@cs.cornell.edu}
\maketitle
\begin{abstract}
We consider a sequence of successively more restrictive definitions of
abstraction for causal 
models, starting with a notion introduced
by Rubenstein et al. \shortcite{Rub17} called 
\emph{exact transformation} that applies to probabilistic causal
models, moving to a 
notion of \emph{uniform transformation} that applies to deterministic
causal models and does not allow differences to be hidden by the
``right'' choice of distribution, and then to
\emph{abstraction}, where the interventions of interest are
determined by the map from low-level states to high-level states, 
and \emph{strong abstraction}, which takes more
seriously all potential interventions in a model, not just the allowed
interventions.  
We show that procedures for combining micro-variables into 
macro-variables are instances of our notion of strong
abstraction, as are all 
the 
examples considered by Rubenstein et al.
\end{abstract}

\section{Introduction}

\noindent We can and 
typically do analyze problems at different levels of
abstraction.  For example, we can try to understand human behavior by
thinking at the level of neurons firing in the brain or at the level of beliefs,
desires, and intentions.  A political scientist might try to
understand an election in terms of individual voters or
in terms of the behavior
of groups such as midwestern blue-collar workers.  Since, in these
analyses, we are typically interested in the \emph{causal} connections
between variables, it seems reasonable to model the various levels of
abstraction using causal models \cite{Hal48,pearl:2k}.  The question
then arises whether a high-level ``macro'' causal model (e.g., one
that considers beliefs, desires, and intentions) is a faithful
abstraction of a low-level ``micro'' model (e.g., one that describes
things at the neuronal level).  What should this even mean?

Perhaps the most common way to approach the question of abstraction is
to cluster ``micro-variables'' in the low-level model into a single
``macro-variable'' in the high-level model \cite{CEP15,CEP16,IS94}.
Of course, one has to be careful to do this in a way that preserves the causal
relationships in the low-level model.  For example, we do not want to
cluster variables $X$, $Y$, and $Z$ into a single variable $X+Y+Z$ if
different settings $(x,y,z)$  
and $(x',y',z')$ such that $x+y+z=x'+y'+z'$ lead to different outcomes.
Rubenstein et al. \shortcite{Rub17} (RW$^+$ from now on) provided an
arguably more general approach to abstraction.  They defined
a notion of an \emph{exact transformation} between two causal models.
They suggest that if there is an exact transformation $\tau$ from causal model
$M_1$ to $M_2$, then we should think of $M_2$ as an abstraction of
$M_1$, so that $M_2$ is the high-level model and $M_1$ is the low-level model.

Abstraction almost by definition involves ignoring inessential
differences.  So it seems that RW$^+$ would want to claim that if there exists an
exact transformation from $M_1$ and $M_2$, then $M_2$ and $M_1$
are the same, except for ``inessential differences''.  This leads to
the obvious question: what counts as an inessential difference?  Of
course, this is to some extent in the eye of the beholder, and may well
depend on the application.  Nevertheless we claim that the notion of
``inessential difference'' implicitly encoded in the definition of exact
transformation is far too broad.  As we show by example, there are
models that we would view as significantly different that are related
by exact transformations. There are 
two reasons for this.  The first is that, 
because RW$^+$ consider probabilistic causal models, some differences
that are intuitively significant are overlooked by considering just the right
distributions.  
Second, besides a function that maps low-level states to high-level
states, RW$^+$ define a separate mapping of interventions that can 
mask what we view as essential differences between the interventions
allowed at the low level and the high level.

In this paper, we consider a sequence of successively more restrictive
definitions of abstraction,
starting with the RW$^+$ notion of exact transformation, moving to a
notion of \emph{uniform transformation} that applies to deterministic
causal models and does not allow differences to be hidden by the
``right'' choice of distribution, and then to
\emph{abstraction}, where the 
mapping between the interventions is determined by the mapping from
low-level states to high-level states,  
and \emph{strong abstraction}, which takes more
seriously all potential interventions in a model, not just the allowed
interventions.  
Finally, we define \emph{constructive abstraction}, which is the special
case of strong abstraction where the mapping from low-level
states to high-level states 
partitions the low-level variables 
and maps each cell to a unique high-level variable. 
As we show, procedures for combining micro-variables into
macro-variables are instances of 
constructive abstraction, as are all the other
examples considered by RW$^+$.
While we view constructive abstraction as the notion that is likely
to be the most useful in practice, as we show by example, the weaker
notions of strong abstraction and abstraction are of interest as well. 

Not surprisingly, the idea 
of abstracting complicated low-level
models to simpler high-level models that in some sense act the same
way has also been considered in other 
settings; see, for example, \cite{BGL17}.  While we are trying to
capture these intuitions as well, considering 
a setting that involves causality adds new subtleties.  

\section{Probabilistic causal models: a review}
In this section we review the definition of causal models.
Much of the discussion is  taken from \cite{Hal48}.

\dfn
A signature $\cal S$ is a tuple $(\U,\V,\R)$, where $\U$
is a set of \emph{exogenous} variables (intuitively, variables that represent
factors outside the control of the model), $\V$ is a set 
of \emph{endogenous} variables (intuitively, variables whose values
are ultimately determined by the values of the endogenous variables),
and $\R$, a function that associates with every variable $Y \in  
\U \union \V$ a nonempty set $\R(Y)$ of possible values for $Y$
(i.e., the set of values over which $Y$ {\em ranges}).
If $\vec{X} = (X_1, \ldots, X_n)$, $\R(\vec{X})$ denotes the
crossproduct $\R(X_1) \times \cdots \times \R(X_n)$.
\edfn

\dfn
A \emph{basic causal model} $M$ is a pair $(\cal S,\F)$, 
where $\cal S$ is a signature and
$\F$ defines a  function that associates with each endogenous
variable $X$ a \emph{structural equation} $F_X$ giving the value of
$X$ in terms of the 
values of other endogenous and exogenous variables (discussed in more
detail below).
A \emph{causal model} $M$ is a tuple $(\cal S,\F,\I)$, where $(\cal S,\F)$ is
a basic causal model and $\I$ is a set
of \emph{allowed interventions} (also discussed in more detail below).
\edfn

Formally, the equation $F_X$ maps $\R(\U \union \V - \{X\})$  to $\R(X)$,
so $F_X$ determines the value of $X$, 
given the values of all the other variables in $\U \union \V$.  
Note that there are no functions associated with exogenous variables;
their values are determined outside the model.  We call a setting
$\vec{u}$ of values of exogenous variables a \emph{context}.

The value of $X$ may depend on the values of only a few other variables.  
$Y$ \emph{depends on $X$ in context $\vec{u}$}
if there is some setting of the endogenous variables  
other than $X$ and $Y$ such that 
if the exogenous variables have value $\vec{u}$, then varying the value
of $X$ in that 
context results in a variation in the value of $Y$; that 
is, there is a setting $\vec{z}$ of the endogenous variables other than $X$ and
$Y$ and values $x$ and $x'$ of $X$ such that $F_Y(x,\vec{z},\vec{u}) \ne
F_Y(x',\vec{z},\vec{u})$.   

In this paper we restrict attention to \emph{recursive} (or
\emph{acyclic}) models, that is, models where, for each context
$\vec{u}$, there is a partial order
%joe3: one more typo.  \prec is not a partial order, because it's not
%reflexive.  So I really do mean a partial order \preceq_u such that
%if X \prec_u Y (where X \prec_u Y if X \preceq_u Y and X \ne Y).
%$\prec_{\vec{u}}$ on variables such that if $X \prec_{\vec{u}} Y$,
%joe3: moreover, the implication is backwards
%$\preceq_{\vec{u}}$ on variables such that if $X \prec_{\vec{u}} Y$,
%then $Y$ depends on $X$ in context $\vec{u}$
$\preceq_{\vec{u}}$ on variables such that if $Y$ depends on $X$ in
context $\vec{u}$, then $X \prec_{\vec{u}} Y$. 
In a recursive model, given a context $\vec{u}$, 
the values of all the remaining variables are determined (we can just
solve for the value of the variables in the order given
by $\prec_{\vec{u}}$).  
A model is \emph{strongly} recursive if the
%joe3
%partial order $\prec_{\vec{u}}$ is independent of $\vec{u}$; that is,
%there is a partial order $\prec$ such that $\prec = \prec_{\vec{u}}$ for all
partial order $\preceq_{\vec{u}}$ is independent of $\vec{u}$; that is,
there is a partial order $\preceq$ such that $\preceq
= \preceq_{\vec{u}}$ for all 
contexts $\vec{u}$.  In a strongly recursive model, we often write the
equation for an endogenous variable as $X = f(\vec{Y})$; this denotes
that the value of $X$ depends only on the values of the variables in
$\vec{Y}$, and the connection is given by $f$.  For example, we might
%joe4
%have $X = Y+U$.
have $X = Y+U$.%
\footnote{RW$^+$ do not restrict to acyclic models.  Rather, they make
  the weaker restriction that, for every setting of the causal
  variables, with probability 1, there is a unique solution to the
  equations.  In the deterministic setting, the analogous restriction
  would be to consider causal models where there is a unique solution
  to all equations.  None of our definitions or results changes if we
  allow this more general class of models.  We have restricted to
  recursive models only to simplify the exposition. }
 
An \emph{intervention} has the form $\vec{X} \gets \vec{x}$, where $\vec{X}$ 
is a set of endogenous variables.  Intuitively, this means that the
values of the variables in $\vec{X}$ are set to $\vec{x}$.  
The structural equations define what happens in the presence of external
interventions.  Setting the value of some variables $\vec{X}$ to
$\vec{x}$ in a causal 
model $M = (\cal S,\F)$ results in a new causal model, denoted $M_{\vec{X}
\gets \vec{x}}$, which is identical to $M$, except that $\F$ is
replaced by $\F^{\vec{X} \gets \vec{x}}$: for each variable $Y \notin
  \vec{X}$, $F^{\vec{X} \gets \vec{x}}_Y = F_Y$ (i.e., the equation
  for $Y$ is unchanged), while for
each $X'$ in $\vec{X}$, the equation $F_{X'}$ for is replaced by $X' = x'$
(where $x'$ is the value in $\vec{x}$ corresponding to $\vec{x}$).

The set $\I$ of interventions can be viewed as the set of interventions
that we care about
for some reason or other. For example, it might consist of the
interventions that involve variables and values that are under our
control.  
In \cite{HP01b,Hal48}, only basic causal models are considered
(and are called causal models).
RW$^+$ added the set of allowed interventions to the model.  We
consider  allowed interventions as well,  since it seems
useful when considering abstractions to describe the set of
interventions of interest.
We sometimes write a causal model $M = (\cal S,\F,\I)$ as $(M',\I)$, where
$M'$ is the basic causal model $(\cal S,\F)$, if we want to emphasize the
role of the set of interventions.

Given a signature $\cal S = (\U,\V,\R)$, a \emph{primitive event} is a
formula of the form $X = x$, for  $X \in \V$ and $x \in \R(X)$.  
A {\em causal formula (over $\cal S$)\/} is one of the form
$[Y_1 \gets y_1, \ldots, Y_k \gets y_k] \phi$,
where
\begin{itemize}
\item
$\phi$ is a Boolean
combination of primitive events,
\item $Y_1, \ldots, Y_k$ are distinct variables in $\V$, and
\item $y_i \in \R(Y_i)$.
\end{itemize}
Such a formula is
abbreviated
as $[\vec{Y} \gets \vec{y}]\phi$.
The special
case where $k=0$
is abbreviated as
$\phi$.
Intuitively,
$[Y_1 \gets y_1, \ldots, Y_k \gets y_k] \phi$ says that
$\phi$ would hold if
$Y_i$ were set to $y_i$, for $i = 1,\ldots,k$.

A causal formula $\psi$ is true or false in a causal model, given a
context. As usual, we write $(M,\vec{u}) \sat \psi$  if the causal
formula $\psi$ is true in causal model $M$ given context $\vec{u}$.
The $\sat$ relation is defined inductively. 
%sander1: rewritten to introduce new notation
%joe1: here I prefer the original.  This doesn't really introduce the
%notation.  
%\commentout{
$(M,\vec{u}) \sat X = x$ if the variable $X$ has value $x$
in the unique (since we are dealing with recursive models) solution
to the equations in $M$ in context $\vec{u}$ (i.e., the unique vector
of values
%joe1
%for the 
%%sander1: exogenous
%endogenous
% variables
that simultaneously satisfies all
equations in $M$ with the variables in $\U$ set to $\vec{u}$).
%}%end
%joe1
%Since we are dealing with recursive models, for each context
% $\vec{u} \in \R(\U)$ there is a unique solution $\vec{v} \in \R(\V)$
% to the equations in $M$. We define $(M,\vec{u}) \sat X = x$ if the
% variable $X$ has value $x$ in this $\vec{v}$. %  
%sander7: why is this uncommented? I added it back.
The truth of conjunctions and negations is defined in the standard way.
Finally, $(M,\vec{u}) \sat [\vec{Y} \gets \vec{y}]\phi$ if 
$(M_{\vec{Y} \gets {y}},\vec{u}) \sat \phi$.

%sander1: added new notation
%joe1
%For ease of notation, sometimes the unique solution $\vec{v}$ to the
To simplify notation, we sometimes write $M(\vec{u})$ to denote the
%joe2:
%values  $\vec{v}$ of the endogenous variables in the (unique) solution
%to the equations in $M$ for a given context $\vec{u}$.
unique element of $\R(\V)$ such that $(M,\vec{u}) \sat \V = \vec{v}$.
%joe1
%is denoted by $M(\vec{u})$.
%joe2
%Likewise, for a given context $\vec{u}$ and
%intervention$\vec{Y} \gets \vec{y}$, we let
%$M(\vec{u},\vec{Y} \gets \vec{y})$  
Similarly, given an intervention $\vec{Y} \gets \vec{y}$, 
$M(\vec{u},\vec{Y} \gets \vec{y})$
%joe1
%denote the unique solution.
%joe2
%denote the values of the endogenous variables in the unique solution
%to the equations.
denotes the unique element of $\R(\V)$ such that $(M,\vec{u}) \sat
[\vec{Y} \gets \vec{y}](\V = \vec{v})$.

A \emph{probabilistic causal model} $M=(\cal S,\F,\I,\Pr)$ is just a
causal model together with a probability $\Pr$ on contexts.  
We often abuse notation slightly and denote the probabilistic causal
model $(\cal S,\F,\I,\Pr)$ as $(M,\Pr)$, where $M$ is the underlying
deterministic causal model $(\cal S,F,\I)$.

RW$^+$ worked with probabilistic causal models, but added one more
feature and made a restrictive assumption.
They consider models $M$ that place a partial order $\prec_M$ on interventions. 
However,  they (and we)
consider only what they call the \emph{natural} partial
order, where $(\vec{X} \gets \vec{x}) \prec_M (\vec{X}' \gets
\vec{x}')$ if $\vec{X}$ is a subset of $\vec{X}'$ and $\vec{x}$ is the
corresponding subset of $\vec{x}'$, so we do not explicitly introduce
the partial order as a component of the model here.
In addition, RW$^+$ assume that for each endogenous
variable $X$, there is a unique exogenous variable $U_X$ such that
$U_X$ is the only exogenous variable on whose value $X$ depends,
and $U_X \ne U_Y$ if $X \ne Y$.
We say that a causal model has \emph{unique exogenous
variables (uev)} if this is the case.

Assuming that a causal model has uev 
makes sense if we think of $U_X$ as the noise variable 
corresponding to $X$.  However, this assumption is not always
appropriate (e.g., if we take the temperature to be exogenous, and
temperature can affect a number of endogenous variables).
Not surprisingly, in (non-probabilistic) causal models, assuming uev
entails a significant loss of
generality. In particular, we cannot express the correlation in values
between two endogenous variables due to being affected by a common
exogenous variable.  However, the uev assumption can be made
essentially without 
loss of generality in probabilistic causal models, as the lemma below shows.

\dfn Two probabilistic causal models
$M = ((\U,\V,\R),\F,\I,\Pr)$ and $M' = ((\U',\V',\R'),\F',\I',\Pr')$ 
are \emph{equivalent},
written $M \sim M'$, if $\V = \V'$, $\R(Y) = \R'(Y)$ for all $Y \in
\V$, $\I = \I'$, and all causal formulas have the same
probability of being true in both $M$ and $M'$; that is, for all
causal formulas $\phi$, we have 
$\Pr(\{\vec{u} \in
\R(\U): (M,\vec{u}) \sat \phi\}) = \Pr'(\{\vec{u}' \in
\R(\U'): (M',\vec{u}') \sat \phi\})$. 
\edfn

\lem\label{lem1} Given a probabilistic causal model
$M$, there is a probabilistic causal model
$M'$ with uev 
such that $M \sim M'$.\footnote{Proofs can be found in the appendix.} 
\elem

All the models that we consider in our examples have uev.  Whatever
problems there are with the RW$^+$ 
notions, they do not arise from the assumption that models have uev.

\section{From exact transformations to abstractions}\label{sec:first}

In this section, we review the RW$^+$ definition, point out some
problems with it, and then consider a sequence of strengthenings of
the definition.

\subsection{Exact transformations}
We need some preliminary definitions.
First observe that, given a probabilistic model $M = ((\U,\V,\R),\F,\I,\Pr)$,
the probability $\Pr$ on $\R(\U)$ can also be viewed as a probability
on $\R(\V)$ (since each context in $\R(\U)$ determines a unique
setting of the variables in $\V$); more precisely,
%sander1: used the new notation
%$$\Pr(\vec{v}) = \Pr(\{\vec{u}: (M,\vec{u}) \sat \V = \vec{v}\}).$$
$$\Pr(\vec{v}) = \Pr(\{\vec{u}: M(\vec{u})= \vec{v}\}).$$
In the sequel, we freely view
$\Pr$ as a distribution on both $\R(\U)$ and $\R(\V)$;
the context should make clear which we intend.
Each intervention $\vec{X} \gets \vec{x}$ also induces a probability
$\Pr^{\vec{X} \gets \vec{x}}$ on $\R(\V)$ in the obvious way:
%sander1: likewise
$${\Pr}^{\vec{X} \gets \vec{x}}(\vec{v})
%= \Pr(\{\vec{u}: (M,\vec{u}) \sat [\vec{X} \gets \vec{x}](\V = \vec{v})\}).$$
= \Pr(\{\vec{u}: M(\vec{u},\vec{X} \gets \vec{x})= \vec{v}\}).$$

One last piece of notation:  We are interested in when a high-level
 model is an abstraction of a low-level model.  In the sequel, we
always use $M_L = ((\U_L,\V_L,\R_L), \F_L,\I_L)$ and $M_H =
((\U_H,\V_H,\R_H), \F_H,\I_H)$ to denote deterministic causal models
(where the $L$ and $H$ stand for \emph{low level} and \emph{high level},
respectively).  
We write $(M,\Pr)$ to denote a probabilistic causal model
that extends $M$.

With this background, we can give the RW$^+$ definition of exact
transformation.  Although the definition was given for probabilistic
causal models that satisfy uev,
it makes sense for arbitrary probabilistic causal models.

\dfn\label{def:exact} If $(M_L,\Pr_L)$ and $(M_H,\Pr_L)$ 
are probabilistic causal models,
$\omega: \I_L \rightarrow \I_H$ is an
order-preserving, surjective mapping
(where $\omega$ is \emph{order-preserving} if, for all interventions 
in $i_1, i_2 \in \I_L$ such that $i_1 \prec_{M_L} i_2$ according  
to the natural order, we have $\omega(i_i) \prec_{M_H} \omega(i_2)$),  
and 
$\tau: \R_L(\V_L)  \rightarrow \R_H(\V_H)$,  then
$(M_H,\Pr_H)$ is an \emph{exact ($\tau$-$\omega$)-transformation} of
$(M_L,\Pr_L)$ if, for every intervention $\do{\vec{Y} \gets
\vec{y}} \in \I_L$, we have 
\begin{equation}\label{eq:exact}
{\Pr}_H^{\omega(\do{\vec{Y} \gets \vec{y}})} =
\tau({\Pr}_L^{\do{\vec{Y} \gets \vec{y}}}),
\end{equation}
where
%joe5
%$\tau(\Pr_L)$ is the ``pushforward'' distribution on $\R(\V_H)$
$\tau(\Pr_L)$ is the ``pushforward'' distribution on $\R_H(\V_H)$
determined by $\tau$ and $\Pr_L$:
$$\tau({\Pr}_L)(\vec{v}_H) = {\Pr}_L(\{\vec{v}_L: \tau(\vec{v}_L) = \vec{v}_H\}).$$
\edfn

The key point here is the requirement
that ${\Pr}_H^{\omega(\do{\vec{Y} \gets \vec{y}})} =
\tau({\Pr}_L^{\do{\vec{Y} \gets \vec{y}}})$.  
Roughly speaking, it says that if you start from the low-level
intervention $\vec{Y} \gets \vec{y}$ and move up to the high-level
model following two distinct routes,
you end up at the same place.  

The first route goes as follows. The intervention
$\vec{Y} \gets \vec{y}$ changes the probability distribution on
low-level outcomes, giving rise to
$\Pr_L^{\do{\vec{Y} \gets \vec{y}}}$ (where an ``outcome'' is a
setting of the endogenous variables). This distribution can be moved
up to the high level by applying $\tau$, giving
$\tau(\Pr_L^{\do{\vec{Y} \gets \vec{y}}})$, which is a distribution on
high-level outcomes.

The second route goes as follows. From the low-level intervention
$\vec{Y} \gets \vec{y}$ we move up to a high-level intervention by
applying $\omega$, giving $\omega(\do{\vec{Y} \gets \vec{y}})$. This
intervention changes the probability distribution on high-level
outcomes, giving rise to $\Pr_H^{\omega(\vec{Y} \gets \vec{y})}$. To
be an exact transformation means that this distribution and the
previous one are identical, for all interventions
$\vec{Y} \gets \vec{y}$. 

Despite all the notation, we hope that the intuition is clear: the
intervention $\do{\vec{Y} \gets \vec{y}}$ 
acts the same way in the low-level model as the intervention
$\omega(\do{\vec{Y}\gets\vec{y}})$ does 
in the high level-model.
(See RW$^+$ for more discussion and intuition.) 
The following example illustrates Definition \ref{def:exact}.

\xam\label{xam:voting}
Consider a simple voting scenario where we have 99 voters who can
either vote for or against a proposition.  The campaign for the
proposition can air some
subset of two advertisements to try to influence how the voters vote.   
The low-level model is
characterized by endogenous variables  $X_i$, $i=1,\ldots,99$, $A_1$,
$A_2$, and $T$, and exogenous variables $U_i$, $i = 1, \ldots, 101$.
$X_i$ denotes voter $i$'s vote,
so $X_i=1$ if voter $i$ votes for the
proposition, and $X_i=0$ if voter $i$ votes against.  $A_i$ denotes
whether add $i$ is run, and $T$ denotes the total number of votes for
the proposition.  
$U_i$ determines how voter $i$ votes as a function of which ads are
run for $i = 1, \ldots, 99$, while $U_{100}$ and $U_{101}$ determine
$A_1$ and $A_2$, respectively.

We can cluster the voters into three groups: $X_1$--$X_{33}$,
$X_{34}$--$X_{66}$, $X_{67}$--$X_{99}$. 
For example, the first group might represent older, wealthy
voters; the second group might represent soccer moms; and the third
group might represent young singles.
Members of the same group are affected by the ads in the same way, meaning
that $Pr(X_i=1 | A_1=a_1 \land A_2=a_2)=Pr(X_j=1 | A_1=a_1 \land A_2=a_2)$ for all $a_1$, $a_2$ 
and all $i$, $j$ that belong to the same group.
The high-level model replaces the variables $X_1, \ldots, X_{99}$ by 
variables $G_1$, $G_2$, and $G_3$, representing the sum of the votes of each group,
it replaces $U_1, \ldots, U_{99}$ by $U_1', U_2', U_3'$, and replaces
$T$ by a binary variable $T'$ that just indicates who won.  The only
interventions allowed in the low-level model are 
interventions to the variables $A_1$ and $A_2$.

We now have an obvious map $\tau$ from $\V_L$ to $\V_H$ that 
maps a low-level state to a high-level state by
taking $G_1$, $G_2$, and $G_3$ to be the 
total vote of the corresponding groups; the map $\omega$ is just the identity.
Given a probability $\Pr_L$ on $\U_L$, there is an obvious probability
$\Pr_H$ on $\U_H$ such that $(M_H,\Pr_H)$ is an exact transformation
of $(M_L,\Pr_L)$.  Note
that it is critical here that we don't allow interventions on the
individual variables $X_i$ at the low level.  For example, it is not clear to what
high-level intervention $\omega$ should map the low-level intervention
$X_3 \gets 1$.   
\exam

RW$^+$ discuss three applications of exact transformations: 
\begin{itemize}
\item a model from which some variables are marginalized;
\item moving from the micro-level to the macro-level by aggregating groups of variables;
\item and moving from a time-evolving dynamical process to a
stationary equilibrium state. 
\end{itemize}
We review the details of their second application here, just to show
how it plays out in our framework. 

\xam\label{xam:rw1}
Let $M_L$ be a causal model with endogenous variables $\vec{X}=\{ X_i:
1 \leq i \leq n \}$ and $\vec{Y}=\{ Y_i: 1 \leq i \leq m \}$,
%joe4: added
exogenous variables $\vec{U}=\{ U_i:
1 \leq i \leq n \}$ and $\vec{V}=\{ V_i: 1 \leq i \leq m \}$,
%joe4: was this typo also in our original version?
%and equations $X_i=U_i$ for $1 \leq i \leq n$ and $Y_i=\sum_{n}^{j=1}
%A_{ij}X_j+V_i$ for $1 \leq i \leq m$, where $A$ is an $m \times n$
and equations $X_i=U_i$ for $1 \leq i \leq n$ and $Y_i=\sum^{n}_{j=1}
a_{ij}X_j+V_i$ for $1 \leq i \leq m$, where $A = (a_{ij})$ is an $m \times n$
matrix, and there exists an $a \in \mathbb{R}$ such that each column
of the matrix $A$ sums to $a$.
Finally, the intervention set is
$\I_L = \{ \emptyset, \vec{X} \gets \vec{x}, \vec{Y} \gets \vec{y},
(\vec{X},\vec{Y}) \gets
(\vec{x},\vec{y}): \vec{x} \in \mathbb{R}^n,\vec{y} \in \mathbb{R}^m \}$.  

Let $M_H$ be a model with endogenous variables $\bar{X}$ and $\bar{Y}$,
%joe4: added
exogenous variables $\bar{U}$ and $\bar{V}$,
equations $\bar{X}=\bar{U}$ and
%sander8: this was wrong, and it's alsp wrong in the original in the
%RW paper. %$\bar{Y}=\frac{a}{m}\bar{X}+\bar{V}$, and intervention set
%joe4
%$\bar{Y}=\frac{a \times n}{m}\bar{X}+\bar{V}$, and intervention set
$\bar{Y}=\frac{a n}{m}\bar{X}+\bar{V}$, and intervention set
$\I_H=  \{ \emptyset, \bar{X} \gets \bar{x}, \bar{Y} \gets \bar{y},
(\bar{X},\bar{Y}) \gets
(\bar{x},\bar{y}): \bar{x} ,\bar{y} \in \mathbb{R} \}$.  
%joe4: removed paragraph break
%
Consider the following transformation that
%joe4
%averages the $X$ and $Y$ variables:
averages $X$ and $Y$:
$$\begin{array}{ll}
%joe5
%\tau:\R(\V_L) \rightarrow \R(\V_H)=\mathbb{R}^2\\
\tau:\R_L(\V_L) \rightarrow \R_H(\V_H)=\mathbb{R}^2\\
%sander8: corrected typo
(\vec{X},\vec{Y}) \rightarrow (\frac{1}{n} \sum_{i=1}^{n}
X_i,\frac{1}{m} \sum_{i=1}^{m} Y_i).
\end{array}$$
%joe4
%
Given a probability $\Pr_{\U_L}$ on $\U_l$ (the contexts in the
low-level model $M_L$), if we take
%sander8: corrected typo
$\Pr_{\bar{U}}=\Pr_{\U_L}(\frac{1}{n} \sum_{i=1}^{n} U_i)$ and
$\Pr_{\bar{V}}=\Pr_{\U_L}(\frac{1}{m} \sum_{i=1}^{m} V_i)$, then $M_H$
is an exact ($\tau$-$\omega$)-transformation of $M_L$ for the obvious
choice of $\omega$. 
\exam

\subsection{Uniform transformations}
As the following 
example shows, much of the work to ensure that a
transformation is an exact transformation can be done by
choosing appropriate distributions $\Pr_L$ and $\Pr_H$.  This leads to
cases where $(M_H,\Pr_H)$ is an exact transformation of
$(M_L,\Pr_L)$ although it is hard to think of $M_H$ as a high-level
abstraction of $M_L$.

\xam\label{xam:notexact}
For $i = \{1,2 \}$, let $M_i$ be a deterministic causal model with signature
$(\U_i,\V_i,\R_i)$; let $\vec{u}_i$ be a fixed context in $M_i$; let
$\vec{v}_i \in \R_i(\V_i)$ be  such that $(M_i,\vec{u}_i) \sat \V_i =
\vec{v}_i$; let $\I_i$ consist only of the empty
intervention;  
let $\Pr_i$ put probability 1 on $\vec{u}_i$;
let $\tau_i$ map all elements of $\R(\V_i)$ to $\vec{v}_{3-i}$;
and let $\omega_i$ be the identity map from $\I_i$ to $\I_{3-i}$.
Clearly $(M_i,\Pr_i)$ is an exact $(\tau_i$-$\omega_i)$-transformation
of $(M_{3-i},\Pr_{3-i})$.
\exam

The fact that each of $M_1$ and $M_2$ is an exact transformation of
the other, despite the fact that the models are completely unrelated,
suggests to us that exact transformations are not capturing the
essence of abstraction.
Roughly speaking, what is happening here is that a high-level model $M_H$ can be
arbitrary in contexts that do not lead to settings $\vec{v}_H$ that
have positive probability for some allowed low-level intervention.  This means
that if there are few allowed low-level interventions or few contexts with
positive probability, then there are very few constraints on $M_H$.
We end up with high-level models $M_H$ that should not (in our view) count as
abstractions of $M_L$.  
We can address this concern by strengthening the
notion of exact 
transformation to 
require it to hold for {\em all} distributions $\Pr_L$. 

\dfn\label{def:uniform} If $M_L$ and $M_H$ are deterministic causal models,
$\omega$ is an order-preserving, surjective mapping $\omega: \I_L
\rightarrow \I_H$, and $\tau: \R_L(\V_L)  \rightarrow \R_H(\V_H)$,  then
$M_H$ is a \emph{uniform ($\tau$-$\omega$)-transformation} of $M_L$
if, for all $\Pr_L$, there exists $\Pr_H$
such that $(M_H,\Pr_H)$ is an exact ($\tau$-$\omega$)-transformation of
$(M_L,\Pr_L)$.   
\edfn

As we pointed out earlier, since RW$^+$ assume uev, the probability
distribution in general might do a lot of work to capture
correlations between values of endogenous variables.  It makes sense
to consider arbitrary distributions if we drop the uev assumption (as
in fact we do).  

In Example~\ref{xam:notexact}
it is easy to see that neither $M_1$ nor $M_2$ is a uniform
transformation of the other.
On the other hand, in Example~\ref{xam:voting}, we do have a uniform
transformation.  

Considering uniform
transformations has other nice features.  For one thing, it
allows us to derive from $\tau$ a mapping $\tau_{\U}$ from
%joe5
%$\R(\U_L)$ to $\R(\U_H)$
$\R_L(\U_L)$ to $\R_H(\U_H)$ 
that ``explains'' how $\Pr_L$ and $\Pr_H$ are related.
More precisely, not only do we know that, for the appropriate
$\omega$, for all distributions $\Pr_L$ there exists $\Pr_H$ such that
$(M_H,\Pr_H)$ is an exact ($\tau$-$\omega$)-transformation
of $(M_L,\Pr_L)$, we can take $\Pr_H$ to be $\tau_{\U}(\Pr_L)$ (i.e.,
the pushforward of $\Pr_L$ under $\tau_{\U}$).  

\pro\label{pro:pushforward} If $M_H$ is a uniform
($\tau$-$\omega$)-transformation of $M_L$
%joe4: added, to take of Marc's concern
and $\I_L$ is countable,
then there 
%joe5
%exists a function $\tau_{\U}: \R(\U_L) \rightarrow \R(\U_H)$
exists a function $\tau_{\U}: \R_L(\U_L) \rightarrow \R_H(\U_H)$
such that, for
%joe5
%all distributions $\Pr_L$ on $\R(\U_L)$, $(M_H,\tau_{\U}(\Pr_L))$ is an
all distributions $\Pr_L$ on $\R_L(\U_L)$, $(M_H,\tau_{\U}(\Pr_L))$ is an
exact ($\tau$-$\omega$)-transformation of $(M_L,\Pr_L)$.   
\epro

The next result provides a characterization of when $M_H$ is a
uniform ($\tau$-$\omega$)-transformation of $M_L$.

%joe3
%\dfn $\tau': \R(\U_L) \rightarrow \R(\U_H)$
%is \emph{compatible} with $\tau: \R(\V_L) \rightarrow \R(\V_H)$ if,
\dfn $\tau': \R_L(\U_L) \rightarrow \R_H(\U_H)$
is \emph{compatible} with $\tau: \R_L(\V_L) \rightarrow \R_H(\V_H)$ if,
%sander1: rephrased it as discussed
\commentout{
for all $\vec{Y} \gets \vec{y} \in \I_L$, $\vec{u}_L \in
%joe3
%\R(\U_L)$, and $\vec{v}_H \in \R(\V_H)$, there exists
%$\vec{v}_L \in \R(\V_L)$   such that
\R_L(\U_L)$, and $\vec{v}_H \in \R_H(\V_H)$, there exists
$\vec{v}_L \in \R_L(\V_L)$   such that
\begin{equation}\label{eq0}
\begin{array}{l}
\hspace{-.15in} (M_H, \tau'(\vec{u}_L)) \sat
 [\omega(\vec{Y} \gets \vec{y})](\V_H = \vec{v}_H)
\mbox{ iff }\\
\hspace{-.15in} (M_L,\vec{u}_L) \sat [\vec{Y}  \gets \vec{y}](\V_L = \vec{v}_L) \mbox{
and } \tau(\vec{v}_L)=\vec{v}_H. 
\end{array}
\end{equation}
}%end
for all $\vec{Y} \gets \vec{y} \in \I_L$ and $\vec{u}_L \in
%joe1
%\R(\U_L)$ it holds that
%joe3
%\R(\U_L)$,
\R_L(\U_L)$,
$$\tau(M_L(\vec{u}_L,\vec{Y}  \gets \vec{y})) =
%joe1
%M_H(\tau'(\vec{u}_L),\omega(\vec{Y} \gets \vec{y}))$$
M_H(\tau'(\vec{u}_L),\omega(\vec{Y} \gets \vec{y})).$$ 
\edfn

\thm\label{thm:uniformchar} 
Given causal models $M_L$ and $M_H$,
%joe3
%$\tau: \R(\V_L) \rightarrow \R(\V_H)$, and an order-preserving
$\tau: \R_L(\V_L) \rightarrow \R_H(\V_H)$, and an order-preserving
surjective function $\omega: \I(\V_L) \rightarrow \I(\V_H)$, 
the following are equivalent: 
\begin{itemize}
  \item[(a)] $M_H$ is a uniform ($\tau$-$\omega$)-transformation of
    $M_L$;
\item[(b)] there exists a function
%joe3
%$\tau_{\U}: \R(\U_L) \rightarrow \R(\U_H)$
$\tau_{\U}: \R_L(\U_L) \rightarrow \R_H(\U_H)$ 
compatible with $\tau$.
\end{itemize}
\ethm

It is easy to check that uniform transformations are closed under
composition.
\thm\label{thm:composition} If $M_H$ is a uniform
($\tau_1$-$\omega_1$)-transformation of 
$M_I$ and $M_I$ is a uniform  ($\tau_2$-$\omega_2$)-transformation of
$M_L$, then $M_H$ is a 
uniform $((\tau_2 \circ \tau_1)$-$(\omega_2 \circ \omega_1))$-transformation
of $M_L$. 
\ethm

\subsection{Abstraction}

Although the notion of a uniform transformation deals with some of the
problems we see with the RW$^+$ notion of exact transformation, it does
not deal with all of them, as the following two examples show.

\xam\label{xam:notabstraction} Let $M_1$ and $M_2$ be deterministic
causal models, both with endogenous 
binary variables $X_1$ and $X_2$  and corresponding 
%joe6
%binary exogenous variables $U_1$ and $U_2$.
binary exogenous variables $U_1$ and $U_2$.%
\footnote{A variable is binary if its range is $\{0,1\}$.}
In $M_1$, the
equations are $X_1 =  U_1$ and $X_2 = X_1$. ($U_2$ plays no role in
the equations in $M_1$.  We added it just to make $M_2$ a model that has
uev and thus show that having uev is not an issue here.)  In $M_2$, the equations are $X_1 = U_1$ and $X_2 = U_2$.  The
only allowed interventions in $M_1$ are $X_1 \gets x_1$, for $x_1 \in
\{0,1\}$; the only allowed 
interventions in $M_2$ are $(X_1,X_2) \gets (x_1, x_1)$, for $x_1 \in
\{0,1\}$.  It is easy to see that $M_1$ is a
uniform transformation of $M_2$ and that $M_2$ is a uniform transformation of
$M_1$.
If $\tau_{ij}$ and $\omega_{ij}$ are the maps showing that $M_j$ is a
uniform transformation of $M_i$, then we can take both $\tau_{12}$ and
$\tau_{21}$ to be the identity, $\omega_{12}$ maps $X_1 \gets x_1$
to $(X_1,X_2) \gets (x_1,x_1)$, while $\omega_{21}$ maps
$(X_1,X_2) \gets (x_1,x_1)$ to $X_1 \gets x_1$.  But this does not
match our intuition that if $M_H$ is 
an abstraction of $M_L$, then $M_H$ is a higher-level description of the
situation than $M_L$.  Whatever ``higher-level description'' means, we
would expect that if $M_L$ and $M_H$ are different, then we
should not have 
$M_L$ and $M_H$ being abstractions of each other.
\exam

What is the problem here? If we just focus on these sets of allowed
interventions,
then there is in fact no problem. $M_1$ and $M_2$ do, in a sense, work
the same way as far as these allowed  
interventions go.
However, the mappings $\omega_{12}$ and $\omega_{21}$ seem to be in
conflict with taking $\tau_{12}$ and $\tau_{21}$ to be the identity. Given
that $\tau_{ij}$ is the identity mapping, we would expect $\omega_{ij}$ to also
be the identity mapping. Why should $\omega_{12}$ map
$X_1 \gets 0$ to something other than $X_1 \gets 0$ here?
It is easy to see that if we take $\omega_{12}$ to also be the
identity mapping then the problem disappears, as we no longer have
uniform transformations between these two models. More generally, we
define below a natural way in which
a mapping $\tau$ on states induces a mapping $\omega$ on allowed
interventions. But even when $\omega$ is well-behaved there exist
counterintuitive examples of uniform transformations. 

\xam\label{xam:new} Given a model $M_3$, let 
$M_4$
be a model that is like $M_3$ except 
that $\U_3$ has a new exogenous binary variable $U^*$
and a new binary endogenous variable $X^*$.   Modify the equations in $M_3$
so that $U^*$ is the only parent of $X^*$, but $X^*$ is the parent of
every other endogenous variable in $M_4$ (and thus of every
endogenous variable in $M_3$).  Take $\I_4^* = \I_3$.
If $X^*=1$, then all equations
in $M_4$ are identical to those in $M_3$.
However, if $X^*=0$, then all equations behave in some arbitrary way
(the exact way they behave is irrelevant). Define
%joe3
%$\tau: \R(\V_3) \rightarrow \R(\V_4)$ by taking
$\tau: \R_3(\V_3) \rightarrow \R_4(\V_4)$ by taking
$\tau(\vec{v}_L) = (\vec{v}_L,X^*=1)$.  
We claim that $M_4$ is a uniform
($\tau$-$\omega$)-transformation of $M_L$, where $\omega$ is the
%joe3
%identity.  Given a distribution $\Pr_3$ on $\R(\U_3)$,
identity.  Given a distribution $\Pr_3$ on $\R_3(\U_3)$, 
define $\Pr_4$ so that its marginal on the variables in $\U_3$ is $\Pr_3$
and $\Pr_4(U^*=0) = 0$.
It is easy to see that $(M_4,\Pr_4)$ is an exact
($\tau$-$\omega$)-transformation of $(M_3,\Pr_3)$, regardless of how 
the equations in $M_4$ are defined if $X^*=1$.
\exam

What goes wrong in this example is that the high level is more
detailed than the low level, contrary to what one expects of an
abstraction. Concretely, introducing the extra variable $X^*$ allows
$M_4$ to capture a whole range of possibilities that have no
counterpart whatsoever in $M_3$. That doesn't sound right (at least to
us). We can fix
this by simply demanding that our abstraction function $\tau$ be
surjective. 

Combining both observations, we define a natural way in which an
abstraction function $\tau$
determines which sets of interventions should be allowed at the low
level and the high level, and the mapping $\omega_{\tau}$ between them.  

\dfn\label{def:induce}
Given a set $\V$ of endogenous variables, $\vec{X} \subseteq \V$, and $\vec{x}
\in \R(\vec{X})$, let 
$$\Rest(\V,\vec{x}) = \{\vec{v} \in \R(\V): \vec{x}
\mbox{ is the restriction of } \vec{v} \mbox{ to } \vec{X}\}.$$
%joe3
%Given $\tau: \R(\V_L) \rightarrow \R(\V_H)$,
Given $\tau: \R_L(\V_L) \rightarrow \R_H(\V_H)$, 
define $\omega_\tau(\vec{X} \gets \vec{x}) = \vec{Y} \gets \vec{y}$ if
%joe3
%there exists $\vec{Y} \subseteq \V_H$ and $\vec{y} \in \R(\vec{Y})$
%joe5
%there exists $\vec{Y} \subseteq \V_H$ and $\vec{y} \in \R_H(\vec{Y})$
%such that $\tau(\Rest(\V_L,\vec{x})) = \Rest(\V_H,\vec{y})$ (as
$\vec{Y} \subseteq \V_H$, $\vec{y} \in \R_H(\vec{Y})$, 
and $\tau(\Rest(\V_L,\vec{x})) = \Rest(\V_H,\vec{y})$ (as
usual, given 
%joe3
%$T \subseteq \R(\V_L)$, we define $\tau(T) = \{\tau(\vec{v}_L):
$T \subseteq \R_L(\V_L)$, we define $\tau(T) = \{\tau(\vec{v}_L):
\vec{v}_L \in T\}$).  
It is easy to see
that, given $\vec{X}$ and $\vec{x}$, there can be at most one such
$\vec{Y}$ and $\vec{y}$.  If there does not exist such a $\vec{Y}$ 
and $\vec{y}$, we take $\omega_\tau(\vec{X} \gets \vec{x})$ to be
undefined.
Let $\I^\tau_L$ be the set of interventions for which
$\omega_\tau$ is defined, and let $\I^\tau_H = \omega_{\tau}(\I^\tau_L)$.  
\edfn

It is straightforward to check that 
in Example~\ref{xam:voting},
$\omega_\tau$ is defined on interventions to $A_1$, $A_2$, and on
these interventions it is the identity (and thus agrees with $\omega$
as defined in 
that example), but it is also defined on simultaneous interventions 
on $X_1-X_{33}$, $X_{34}-X_{66}$, and
$X_{67}-X_{99}$, and on $T$ (as well as combinations of these
interventions).  In Example~\ref{xam:rw1}, the interventions on which
$\omega_\tau$ is defined are precisely those in the set $\I_L$ of that
example; on these interventions, $\omega_\tau = \omega$.  

Note that if  $\tau$ is surjective, then it 
follows that
$\omega_{\tau}(\emptyset)=\emptyset$, 
%joe3
%and for all $\vec{v}_L \in \R(\V_L)$, $\omega_{\tau}(\V_L \gets
and for all $\vec{v}_L \in \R_L(\V_L)$, $\omega_{\tau}(\V_L \gets
  \vec{v}_L)=\V_H \gets \tau(\vec{v}_L)$. 

 \dfn\label{dfn:tau-abstraction}
$(M_H, \I_H)$ is a \emph{$\tau$-abstraction of $(M_L, \I_L)$} if
the following conditions hold:
\begin{itemize}
\item $\tau$ is surjective;
\item there is a surjective function
%joe3
%$\tau_{\U}: \R(\U_L) \rightarrow \R(\U_H)$ compatible with $\tau$;
$\tau_{\U}: \R_L(\U_L) \rightarrow \R_H(\U_H)$ compatible with $\tau$;
\item $\I_H = \omega_{\tau}(\I_L)$.
\ \  \bbox
\end{itemize}
\end{definition}

As intended, Examples~\ref{xam:notabstraction} and \ref{xam:new} are
not $\tau$-abstractions; on the other hand, in Examples~\ref{xam:voting}
and~\ref{xam:rw1}, 
$M_H$ is a $\tau$-abstraction of $M_L$.

%joe6: uniform transformation are also relations between causal models
%Unlike exact and uniform transformations,
Unlike exact transformations,
$\tau$-abstraction is a relation between causal models: the mapping
$\omega$ is determined by $\tau$, and there is no need to specify a
probability distribution. 

%joe6*: added proposition and proof (mainly taken fromthe
%approximation paper, but correcting some typos
\pro If $M_H$ is a $\tau$-abstraction of $M_L$, then $M_H$ is a uniform
$(\tau,\omega_\tau)$-tranformation of $M_L$.  \epro\label{thm:abstuniform}

%sander9: moved the proof to the appendix

We can strengthen the notion of $\tau$-abstraction to define a relation
  on basic causal models, by
  considering the largest possible sets of allowed interventions.
 \dfn\label{def:strong} If $M_H$ and $M_L$ are basic causal models,
 then $M_H$ is a \emph{strong $\tau$-abstraction} of $M_L$ if
$\I_H^{\tau}=\I_H^*$, the set of all high-level interventions, and
$(M_H,\I_H^{\tau})$ is a $\tau$-abstraction 
of $(M_L,\I_L^{\tau})$. 
\edfn

The notion of strong $\tau$-abstraction provides a clean,
powerful relation between basic causal models.
However, there are applications where the two additional
requirements that make an abstraction strong are too much to ask. In
the following example, neither requirement is satisfied.

\xam\label{xam:new3} Consider an object in the
earth's gravitational field. On the low level ($M_L$), there are
three endogenous variables: $V$ (velocity), $H$ (height), and $M$
(mass), and three corresponding exogenous variables, $U_V$, $U_H$, 
and $U_M$.  
The equations in $M_L$ are $V = U_V$, $H = U_H$, and $M = U_M$. The
high level captures the object's current energy. $M_H$ contains
endogenous variables $K$ (kinetic energy) and $P$ (potential energy),
and two corresponding exogenous variables, $U_K$ and $U_P$. The
equations in $M_H$ are $K = U_K$ and $P = U_P$. 
%joe3
%We define $\tau: \R(\V_L) \rightarrow \R(\V_H)$
We define $\tau: \R_L(\V_L) \rightarrow \R_H(\V_H)$  
using the standard equations for kinetic energy and gravitational
potential energy, so 
$\tau(v,h,m) = (\frac{1}{2} m v ^2, 9.81 m h)$.
%joe3
%It is easy to see that $\tau$ is a surjection onto $\R(\V_H)$.
It is easy to see that $\tau$ is a surjection onto $\R_H(\V_H)$.
We claim that $M_H$ is not a strong $\tau$-abstraction of $M_L$.
To see why, consider
interventions of the form $M \gets m$ for $m > 0$. Applying
Definition \ref{def:induce}, we get that $\omega_{\tau}(M \gets m)
= \emptyset$, since $\tau(\Rest(\V_L,m)) = \V_H$; by choosing $v$ and
$h$ appropriately, we can still get all values in $\V_H$, as long as
$m > 0$.  We also clearly have that $\omega_\tau$ maps the empty intervention
in $M_L$ to the empty intervention in $M_H$.  With this, we can
already show that 
$(M_H,\I_H^{\tau})$ is not a uniform $(\tau,\omega_\tau)$-
transformation of $(M_L,\I_L^{\tau})$.  
Suppose that $\Pr_L$ is a probability on
$\U_L$ that puts probability 1 on $(1,1,1)$.  For condition
(\ref{eq:exact}) in Definition~\ref{def:exact} to
hold for the  intervention $M \gets m$, 
the probability $\Pr_H$ on $\U_H$ must put probability 1 on
$(.5m,9.81m)$.  But (\ref{eq:exact}) must hold for all choices of $m$.
This is clearly impossible.

Although $M_H$ is not a strong $\tau$-abstraction of $M_L$, we can easily
construct a sensible and useful 
 $\tau$-abstraction between these models by simply not allowing
interventions of the form $M \gets m$ in the low-level model.
Concretely, if we define $\I_L$ as containing the
empty intervention and all interventions of the form $(V,H,M) \gets
(v,h,m)$, then 
$\omega_{\tau}$ maps this to the set $\I_H$
that contains the empty intervention and all interventions of the form
$(K,P) \gets (k,p)$.  
\exam

As the following example shows, there also exist interesting cases 
where only the first requirement of Definition \ref{def:strong} is not
satisfied. Roughly speaking,
this is because some high-level variables are not logically
independent, so not all high-level interventions are meaningful.

\xam\label{xam:pixel}  Suppose that we have a $100 \times 100$
grid of pixels, each of which can be black or white. In the low-level
model, we have 10,000 endogenous 
variables $X_{ij}$, for $1 \le i, j \le 100$, and 10,000 corresponding
exogenous variables $U_{ij}$  for $1 \le i, j \le 100$, with the
obvious equations $X_{ij} = U_{ij}$.  We would expect there to be
other variables that are affected by the $X_{ij}$s (e.g., what a viewer
perceives), but for ease of exposition, we ignore these other
variables in this example and focus only on the $X_{ij}$ variables.
Suppose that all we care about is how many of the pixels 
in the upper half of the grid are black and how many pixels in the
left half of the grid are black. Thus, in the high-level model, we
have variables $\UH$ and $\LH$ whose range is $\{0,\ldots, 5000\}$.
Because of the dependencies between $\UH$ and $\LH$, there is a single
exogenous variable that determines their values, which are pairs
$(m,m')$ such that $0 \le  m, m' \le 5,000$ and $|m-m'| \le 2,500$.
Now we have an obvious map $\tau$ from low-level states to high-level states.
We claim that $(M_H,\I_H^{\tau})$ is a $\tau$-abstraction of $(M_L,\I_L^{\tau})$, where 
$\I_L^{\tau}$
consists of the empty intervention and interventions that
simultaneously set all the variables in the upper half and left half
(i.e., all variables $X_{ij}$ with $1 \le i \le 50$ or $51 \le j \le
100$) and an arbitrary 
subset of the variables in the bottom right.  Given a nonempty
intervention $\vec{X} \gets \vec{x}$ of this form,
$\omega_\tau(\vec{X} \gets \vec{x}) = (\UH \gets
m, \LH \gets m')$, 
where $m$ is the number of $X_{ij}$ variables set to 1 with $1 \le i \le 50$
and $m'$ is the number of $X_{ij}$ variables set to 1 with $51 \le
j \le 100$; how the variables in the bottom right
are set in $\vec{X} \gets \vec{x}$ is irrelevant.  Thus, 
$\I_H^{\tau}$ consists of interventions of the form $(\UH \gets
m, \LH \gets m')$, where $1 \le m, m' \le 5000$ and $|m-m'| \le 2500$.  
It is straightforward to check that there is no low-level intervention
$\vec{X} \gets \vec{x}$ such that $\omega_\tau(\vec{X} \gets \vec{x}) = \UH \gets m$.  For suppose that 
$\omega_\tau(\vec{X} \gets \vec{x}) = \UH \gets m$.  Then 
$\tau(\Rest(\V_L,\vec{x})) = \{(m,m'): 1 \le m' \le 5000\}$.  This means
that $(m,m') \in \tau(\Rest(\V_L,\vec{x}))$ for some $m'$ such that
$|m - m'| > 2500$, which is a contradiction.  A similar argument
shows that no intervention of the form $\LH \gets m'$ can be in 
$\I_H^{\tau}$.
It is straightforward to check that 
$(M_H,\I_H^{\tau})$  is a uniform
$(\tau,\omega_\tau)$-transformation of 
$(M_L,\I_L^{\tau})$, so 
$(M_H,\I_H^{\tau})$ is a $\tau$-abstraction of $(M_L,\I_L^{\tau})$, 
however it is clearly not a strong
$\tau$-abstraction of $M_L$.
\exam

The problem here is that although $M_H$ has variables $\UH$ and $\LH$,
we can only intervene on them simultaneously.  It may make sense to
consider such interventions if we want a visual effect that depends on
both the number of black pixels in the upper half and the number of
black pixels in the left half.  But it is worth noting that if we
consider a high-level model $M_H'$ with only 
a single variable $\ULH$ that counts the number
of pixels that are black in the upper half and the left half
altogether, then $M_H'$ is a strong $\tau$-abstraction of $M_L$
with the obvious map $\tau$.

\fullv{Lastly, we present an example where the second requirement of
Definition \ref{def:strong} is not satisfied.}
\shortv{The full paper (posted on arxiv)
gives an example where the second requirement of
Definition \ref{def:strong} is not satisfied.}

\fullv{
\xam\label{xam:new2} Let $M_1$ be the model with the following
equations, $X_1=U_1$, $X_2=U_2$, and $X_3=U_3$,  
where all variables are binary. Let $M_2$ be a model with
two equations $Y_1 = W_1$ and $Y_2 = W_2$,  
%joe3
%where again all variables are binary. Define $\tau: \R(\V_1) 
%\rightarrow \R(\V_2)$ by taking $\tau(x_1,x_2,x_3) = (x_1 \lor x_3,
where again all variables are binary. Define $\tau: \R_1(\V_1) 
\rightarrow \R_2(\V_2)$ by taking $\tau(x_1,x_2,x_3) = (x_1 \lor x_3,
x_2 \lor x_3)$.   
We leave it to the reader to verify that $\I_2^{\tau}=\I_2^*$ and $\I_1^{\tau}=\I_1^* \setminus \{ (X_1,X_2) \gets (0,0) \}$. 

We claim that $(M_2,\I_2^{\tau})$ is not a $\tau$-abstraction of
$(M_1,\I_1^{\tau})$, and thus that $M_2$ is not a strong
$\tau$-abstraction of $M_1$. To see why, note that both the
low-level intervention $X_3 \gets 0$ and the empty intervention are
mapped to the empty intervention by $\omega_{\tau}$.
Now suppose that we have a prior $\Pr_L$ such that
$\Pr_L(U_1=0)= a$, $\Pr_L(U_2=0) = b$, $\Pr_L(U_3=0) = c$ where $c<1$,
and $U_1$, 
$U_2$, $U_3$ are independent. Thus, $\Pr_L(0,0,0) = abc$.
Applying condition (\ref{eq:exact}) in Definition~\ref{def:exact} to
the empty intervention, we must have 
$\Pr_H(0,0) = abc$. On the other hand, $\Pr_L^{X_3 \gets 0}(0,0,0) =
ab$.  Since $\omega_{\tau}(X_3 \gets 0)$ is the empty intervention, we must also
have $\Pr_H(0,0) = ab$.  This can't happen unless $c=1$,
so $(M_2,\I_2^\tau)$ is not a uniform $(\tau-\omega_\tau)$-transformation of
$(M_,\I_1^\tau)$, and thus cannot be a $\tau$-abstraction.

Now define $\I_L$ so that the only allowable low-level interventions
are ones where $X_3 \gets 0$ 
(i.e., we allow all interventions of the form $\vec{X} \gets \vec{x}$
where $X_3 \gets 0$ is one of the components of
$\vec{X} \gets \vec{x}$;  in particular, we do not allow the empty low-level
intervention). Then clearly $(M_2,\I_2^{\tau})$ is a
$\tau$-abstraction of $(M_1,\I_1)$, where we have that
$\I_2^{\tau}=\I_2^*$. 
\exam
}

\subsection{From micro-variables to macro-variables}
Roughly speaking, the intuition for clustering micro-variables into
macro-variables  is that in the
high-level model, one variable captures the effect of a number of
variables in the low-level model.  This makes sense only if the low-level
variables that are being clustered together ``work 
the same way'' as far as the allowable interventions go.  
The following definition makes this precise.

\dfn $M_H$ is
a \emph{constructive} $\tau$-abstraction 
of $M_L$ if $M_H$ is a 
strong $\tau$-abstraction of $M_L$ and, if
$\V_H = \{Y_1, \ldots, Y_n\}$, then there exists 
a partition $P = \{\vec{Z}_1, \ldots, \vec{Z}_{n+1}\}$ of $\V_L$,
where $\vec{Z}_1, \ldots, \vec{Z}_n$ are nonempty,
and
%joe3
%mappings $\tau_i: \R(\vec{Z}_i) \rightarrow \R(Y_i)$ for $i=1, \ldots, n$
mappings $\tau_i: \R_L(\vec{Z}_i) \rightarrow \R_H(Y_i)$ for $i=1, \ldots, n$
such that $\tau = (\tau_1, \ldots, \tau_n)$; that is,
$\tau(\vec{v}_L) = \tau_1(\vec{z}_1)\cdot \ldots \cdot
\tau_n(\vec{z}_n)$, where $\vec{z}_i$ is the projection of $\vec{v}_L$
onto the variables in $\vec{Z}_i$, and $\cdot$ is the concatenation
operator on sequences.
$M_H$ is a \emph{constructive
  abstraction} of $M_L$ if it is a \emph{constructive
  $\tau$-abstraction} of $M_L$ for some $\tau$.
\edfn

In this definition, we can think of each $\vec{Z}_i$ as describing a
set of microvariables that are mapped to a single macrovariable $Y_i$.
The variables in $\vec{Z}_{n+1}$ (which might be empty) are ones that are
marginalized away.

By definition, every constructive $\tau$-abstraction is a strong
$\tau$-abstraction. We conjecture that a converse to this also holds: 
that is, if $M_H$ is a strong $\tau$-abstraction of $M_L$,
that perhaps satisfies a few minor technical conditions, then it will in
fact be a constructive $\tau$-abstraction of $M_L$.
However, we have not proved this result yet. 

We suspect that constructive $\tau$-abstractions are the notion of
abstraction that will arise most often in practice.  All three of the examples
discussed by RW$^+$ (one of which is Example~\ref{xam:rw1}) are
constructive abstractions. We can easily extend
Example~\ref{xam:voting} 
by adding low-level and high-level interventions 
to make it a constructive abstraction as well.

\section{Discussion and Conclusions}

We believe that getting a good notion of abstraction will be critical
in allowing modelers to think at a high level while still being faithful
to a more detailed model.  As the analysis of this paper shows, there are
different notions of abstraction, that relate causal models at
different levels of detail.
For example,
$\tau$-abstraction is a relation between
basic causal models, while a uniform $(\tau,\omega)$-
transformation relates causal models, and RW$^+$'s notion of
exact transformation relates probabilistic causal models.
Although our final notion of constructive abstraction is the cleanest
and arguably  easiest to use, we believe that there exist applications
for which the weaker abstraction relations are more appropriate. More
work needs to be done to understand which abstraction relation  
is most suitable for a given application. We hope that the definitions proposed here will help
clarify the relevant issues. They should also shed light on some of the recent discussions of
higher-level causation in communities ranging from physics to
philosophy (see, e.g., \cite{FG17,HAT13}.

In fact, we see the current paper as laying the formal groundwork for
several interesting topics that we intend to explore in future
work. First, we hope to generalize the abstraction relation to a
notion of {\em approximate abstraction}, given that in most real-life
settings the mappings between different levels are only approximately
correct.
Second, our framework makes it possible to explore whether the notion of actual causation could be applied across causal
models, rather than merely within a single causal model.
For example, it seems to be useful to think of an event in a
low-level model as causing an event in a high-level model.
Third, abstracting causal models of large complexity into simpler
causal models with only a few variables is of direct relevance to the
increasing demand for explainable AI, for in many
situations the problem lies not in the fact that no causal model is
available, but in the fact that the only available model is too
complicated for humans to understand. 

\paragraph{Acknowledgments:}
Halpern was supported in part by NSF 
grants IIS-1703846 and IIS-1718108, 
ARO grant W911NF-17-1-0592, and a grant from the Open Philanthropy
project. 
Beckers was supported by the grant ERC-2013- CoG project REINS, nr. 616512.
Some of this work was done while Beckers was a postdoc at Cornell
University, supported by the Belgian American Educational Foundation.
%joe4
%We thank Frederick Eberhardt and the reviewers of the paper
We thank Frederick Eberhardt, Marc Denecker, and the reviewers of the paper
for many useful comments.

\appendix
\section{Appendix: Proofs}

{\bf Proof of Lemma~\ref{lem1}:}
Let $M = ((\U,\V,\R),\F,\I,\Pr)$ and define
$M' = ((\U',\V,\R'),\F',\I,\Pr')$ as follows.
$\U'$ has one exogenous variable for each endogenous variable in $\V$.
Taking $\V = \{Y_1, \ldots, Y_n\}$, we 
%joe3
%take $U_i$ to be the exogenous variable corresponding to $Y_i$.
take $U_i'$ to be the exogenous variable corresponding to $Y_i$.
Let $\U' = \{U_1', \ldots, U_n'\}$.
We take $\R(U_i') = \R(\U)$ for $i = 1, \ldots, n$
(so the set of possible values for each variable $U_i'$ is the set of
all contexts in $M$).
If $\vec{z} \in
\R(\V - \{Y_i\})$, we define 
$F'_{Y_i}(\vec{z},\vec{u}_1, \ldots, \vec{u}_n) =
F_{Y_i}(\vec{z},\vec{u}_i)$.
(Note that here $\vec{u}_i \in \R(\U) = \R(U_i')$.)
Thus, it is clear that the only
exogenous variable that the value of
$Y_i$ in $M'$ depends on is $U_i'$, so $M'$ has uev, as desired.
$\Pr'$ places
probability 0 on a context $(\vec{u}_1,\ldots, \vec{u}_n)$ unless
$\vec{u}_1 = \ldots = \vec{u}_n$, and 
$\Pr'(\vec{u},\ldots, \vec{u}) =
\Pr(\vec{u})$.   It is almost immediate that, with these choices, $M \sim M'$.
\eprf

{\bf Proof of Proposition~\ref{pro:pushforward}:} Suppose that $M_H$
is a uniform ($\tau$-$\omega$)-transformation of $M_L$. 
%sander6: new notation implemented
%For each context $\vec{u}_L \in \R(\U_L)$ and intervention $\vec{Y}
%\gets \vec{y} \in \I_L$, let $\vec{v}_L^{\vec{u}_L,\vec{Y} \gets
 % \vec{y}}$ denote the unique element of $\R(\V_L)$ such that 
%$(M_L,\vec{u}_L) \sat [\vec{Y} \gets \vec{y}](\V_L =
  %\vec{v}_L^{\vec{u}_L,\vec{Y} \gets \vec{y} })$.  We
  %can similarly define $\vec{v}_H^{\vec{u}_H,\vec{Y}' \gets \vec{y}'}$
    %for each context $\vec{u}_H$ in $M_H$ and intervention
    %$\vec{Y}' \gets \vec{y}' \in \I_H$. 
    Say that 
    $\vec{u}_L$ and $\vec{u}_H$ \emph{correspond} if
    %$ \vec{v}_H^{\vec{u}_H,\omega(\vec{Y} \gets \vec{y})} =
  %\tau(\vec{v}_L^{\vec{u}_L,\vec{Y} \gets \vec{y}})$ 
  $\tau(M_L(\vec{u}_L,\vec{Y} \gets \vec{y}))=M_H(\vec{u}_H,\omega(\vec{Y} \gets \vec{y}))$
  for all interventions $\vec{Y} \gets \vec{y} \in \I_L$.  

%joe5
%We claim that for all $\vec{u}_L \in \R(\U_L)$, there exists at least
%one $\vec{u}_H \in \R(\U_H)$ that corresponds to $\vec{u}_L$.  To see this,
%fix $\vec{u}_L \in \R(\U_L)$.  Let $\Pr_L$ give $\vec{u}_L$ probability 1.
We claim that for all $\vec{u}_L \in \R_L(\U_L)$, there exists at least
one $\vec{u}_H \in \R_H(\U_H)$ that corresponds to $\vec{u}_L$.  To see this,
fix $\vec{u}_L \in \R_L(\U_L)$.  Let $\Pr_L$ give $\vec{u}_L$ probability 1.
Then for each intervention $\vec{Y} \gets \vec{y} \in \I_L$, the
distribution $\Pr_L^{\vec{Y} \gets \vec{y}}$ gives probability 1 to
%$\vec{v}_L^{\vec{u}_L,\vec{Y} \gets \vec{y}}$.
$M_L(\vec{u}_L,\vec{Y} \gets \vec{y})$.
Let $\Pr_H$ be 
a probability distribution such that
$(M_H,\Pr_H)$ is an exact ($\tau$-$\omega$)-transformation of  $(M_L,\Pr_L)$.
  Since $\tau(\Pr_L^{\vec{Y} \gets \vec{y}})
  =\Pr_H^{\omega(\vec{Y} \gets \vec{y})}$, it follows that 
    $\Pr_H^{\omega(\vec{Y} \gets \vec{y})}$ gives probability 1 to
      %$\tau(\vec{v}_L^{\vec{u}_L,\vec{Y} \gets \vec{y}})$, and hence
      $\tau(M_L(\vec{u}_L,\vec{Y} \gets \vec{y}))$, and hence
      also to the
      set 
      %$\U_H^{\vec{u}_L,\vec{Y} \gets \vec{y}} = \{\vec{u}_H: (M_H,\vec{u}_H)
      %\sat [\omega(\vec{Y} \gets
            % \vec{y})](\V_H=\tau(\vec{v}_L^{\vec{u}_L,\vec{Y}\gets\vec{y}}))\}$.
            $\U_H^{\vec{u}_L,\vec{Y} \gets \vec{y}} = \{\vec{u}_H: M_H(\vec{u}_H,\omega(\vec{Y} \gets
            \vec{y}))=\tau(M_L(\vec{u}_L,\vec{Y} \gets \vec{y}))\}$.
%joe3*: we never made any explicit assumptions about the cardinality
%of \I_L.  I think that I was thinking that all components of models
%are finite, but we never said this.  We should either say so
%explicitly, or else we can assume in the statement of the Proposition 3.6
%that \I_L is countable (since the argument works in that case as well).
%sander8: I overlooked this. I do recall that you made a similar assumption when proving the next result, and that I could drop it because of the new notation. 
%joe4: so how would you like to deal with it?  I made te minimal
%change, but I could go for other solutions
%Since there are only finitely many interventions in $\I_L$,
Since there are only countably many interventions in $\I_L$,
      $\inter_{\vec{Y} \gets \vec{y} \in \I_L}\U_H^{\vec{u}_L,\vec{Y} \gets
  \vec{y}}$ also has probability 1, 
      and thus must be nonempty.  Choose $\vec{u}_H \in
      \inter_{\vec{Y} \gets \vec{y} \in \I_L}\U_H^{\vec{Y} \gets
        \vec{y}}$.  By construction, $\vec{u}_H$ corresponds to
      $\vec{u}_L$.
      
      Define $\tau_{\U}$ by taking $\tau_{\U}(\vec{u}_L) = \vec{u}_H$, where
$\vec{u}_H$ corresponds to $\vec{u}_L$.  (If more than one tuple
$\vec{u}_H$ corresponds to $\vec{u}_L$, then one is chosen
arbitrarily.) It is now straightforward to check that
$(M_H,\tau_{\U}(\Pr_L))$ is an exact ($\tau$-$\omega$)-transformation of
$(M_L,\Pr_L)$.  We leave details to the reader.
\eprf

%sander6: changed proof to match new version
{\bf Proof of Theorem~\ref{thm:uniformchar}:}
To show that (a) implies (b), suppose that $M_H$ is a uniform
($\tau$-$\omega$)-transformation of  $M_L$.   Let $\tau_{\U} :
%joe5
%\R(\U_L) \rightarrow \R(\U_H)$ be
\R_L(\U_L) \rightarrow \R_H(\U_H)$ be
the function guaranteed to exist by Proposition~\ref{pro:pushforward}.
We must show that
for all $\vec{Y} \gets \vec{y} \in \I_L$, $\vec{u}_L \in
%joe5
%\R(\U_L)$, %and $\vec{v}_H \in \R(\V_H)$,
\R_L(\U_L)$, %and $\vec{v}_H \in \R_H(\V_H)$,
%there exists $\vec{v}_L \in \R(\V_L)$ such that 
%(\ref{eq0}) holds.
$$\tau(M_L(\vec{u}_L,\vec{Y}  \gets \vec{y})) =
M_H(\tau_{\U}(\vec{u}_L),\omega(\vec{Y} \gets \vec{y})).$$ 
%Fix $\vec{Y} \gets \vec{y} \in \I_L$, $\vec{u}_L \in
%\R(\U_L)$, and $\vec{v}_H \in \R(\V_H)$.
%sander6: here starts the simplification
%joe5
%Fix $\vec{u}_L \in \R(\U_L)$. From the construction of $\tau_{\U}$ in the
Fix $\vec{u}_L \in \R_L(\U_L)$. From the construction of $\tau_{\U}$ in the
%joe2
%proof of Proposition~\ref{pro:pushforward} it directly follows that
proof of Proposition~\ref{pro:pushforward}, it follows that  
$\vec{u}_L$ and $\tau_{\U}(\vec{u}_L)$ correspond,
%joe2
%which is defined precisely as
which, by definition, means that
$\tau(M_L(\vec{u}_L,\vec{Y} \gets \vec{y}))=M_H(\tau_{\U}(\vec{u}_L),\omega(\vec{Y} \gets \vec{y}))$
for all interventions $\vec{Y} \gets \vec{y} \in \I_L$.   
  %sander6: I started changing the proof step-wise, until I realized that using our new notation one direction of the proof becomes trivial.
\commentout{
Fix $\vec{Y} \gets \vec{y} \in \I_L$ and $\vec{u}_L \in \R(\U_L)$. 
%Choose $\vec{v}_L$ such that
%$(M_L,\vec{u}_L) \sat [Y \gets y](\V_L = \vec{v}_L)$.
%We now show that the ``iff'' of (b) holds, with this choice of
%$\vec{v}_L$.
%Suppose that
%$(M_H, \tau_{\U}(\vec{u}_L)) \sat [\omega(\vec{Y} \gets \vec{y})](\V_H =
  %\vec{v}_H)$.
Choose $\Pr_L$ such that $\Pr_L(\vec{u}_L) = 1$.  By
  Proposition~\ref{pro:pushforward}, $(M_H,\tau_{\U}(\Pr_L))$ is
  %sander6: a uniform
an exact
  ($\tau$-$\omega$)-transformation of $(M_L,\Pr_L)$.  Thus, 
$\tau(\Pr_L^{\vec{Y} \gets \vec{y}})
  =\tau_{\U}(\Pr_L)^{\omega(\vec{Y} \gets \vec{y})}$. 
 By construction,
%  $\Pr_L^{\vec{Y} \gets \vec{y}}(\vec{v}_L) = 1$, so 
  $\Pr_L^{\vec{Y} \gets \vec{y}}(M_L(\vec{u}_L,\vec{Y}  \gets \vec{y})) = 1$, so 
 % $\tau(\Pr_L^{\vec{Y} \gets \vec{y}})(\tau(\vec{v}_L)) = 1$ and 
  $\tau(\Pr_L^{\vec{Y} \gets \vec{y}})(\tau(M_L(\vec{u}_L,\vec{Y}  \gets \vec{y}))) = 1$ and 
hence 
%$\tau_{\U}(\Pr_L)^{\omega(\vec{Y} \gets \vec{y})}(\tau(\vec{v}_L)) = 1$.  
$\tau_{\U}(\Pr_L)^{\omega(\vec{Y} \gets \vec{y})}(\tau(M_L(\vec{u}_L,\vec{Y}  \gets \vec{y}))) = 1$.  
By definition,
$\tau_{\U}(\Pr_L)(\tau(\vec{u}_L)) = 1$;
from the construction of $\tau_{\U}$ in the proof of
Proposition~\ref{pro:pushforward}, 
%sander6: I stopped changing it here, because I then realized the above simplification.
$(M_H,\tau_{\U}(\vec{u}_L)) \sat [\omega(\vec{Y} \gets \vec{y})](\V_H =
\tau(\vec{v}_L))$.  Since, by assumption,
$(M_H,\tau_{\U}(\vec{u}_L)) \sat [\omega(\vec{Y} \gets \vec{y})](\V_H =
\vec{v}_H)$, we must have $\tau(\vec{v}_L) = \vec{v}_H$.

Conversely, if  $\tau(\vec{v}_L) = \vec{v}_H$,
then by the construction in Proposition~\ref{pro:pushforward},
$(M_H,\tau_{\U}(\vec{u}_L)) \sat [\omega(\vec{Y} \gets \vec{y})](\V_H =
\tau(\vec{v}_L))$,
so
$(M_H,\tau_{\U}(\vec{u}_L)) \sat [\omega(\vec{Y} \gets \vec{y})](\V_H
= \vec{v}_H)$.
}%end

To show that (b) implies (a), suppose that
(b) holds.  Given a distribution $\Pr_L$ on
%joe5
%$\R(\U_L)$, let $\Pr_H =
$\R_L(\U_L)$, let $\Pr_H = 
\tau_{\U}(\Pr_L)$.  It suffices to show that $(M_H,\Pr_H)$ is an exact
($\tau$-$\omega$)-transformation of $(M_L,\Pr_L)$.  Thus, we must show
that for every intervention $\vec{Y} \gets
\vec{y} \in \I_L$, we have 
${\Pr}_H^{\omega(\do{\vec{Y} \gets \vec{y}})} =
\tau({\Pr}_L^{\do{\vec{Y} \gets \vec{y}}}).$  
Straightforward computations now show that
%sander6: rewritten
\commentout{
$$\begin{array}{lll}
  &{\Pr}_H^{\omega(\do{\vec{Y} \gets \vec{y}})}(\vec{v}_H) \\
  = & {\Pr}_H(\{\vec{u}_H:
  (M_H,\vec{u}_H) \sat [\omega(\vec{Y} \gets \vec{y})](\V_H =
  \vec{v}_H)\})\\
  %sander6: note that this could already have been simplified, as
  %indicated, and likewise in the following step, which I've left as
  %is since this is all replaced anyway. 
%  = &{\Pr}_L(\{\vec{u}_L: \exists \vec{u}_H(\tau_{\U}(\vec{u}_L) =
%\vec{u}_H \mbox{ and }
%\\&
%\hspace{.3in}
%(M_H,\vec{u}_H) \sat [\omega(\vec{Y} \gets \vec{y})](\V_H =
= &{\Pr}_L(\{\vec{u}_L: (M_H,\tau_{\U}(\vec{u}_L)) \sat [\omega(\vec{Y} \gets \vec{y})](\V_H =
  \vec{v}_H)\})\\
  = &{\Pr}_L(\{\vec{u}_L: \exists \vec{u}_H, \vec{v}_L (\tau_{\U}(\vec{u}_L) =
  \vec{u}_H, \tau(\vec{v}_L) = \vec{v}_H,
\\ & \hspace{.3in} \mbox{ and } 
    (M_L,\vec{u}_L) \sat [\vec{Y} \gets \vec{y}](\V_L = \vec{v}_L))\})\\
= &{\Pr}_L(\{\vec{v}_L: \tau(\vec{v}_L) =
\vec{v}_H  \mbox{ and } \\ &\hspace{.3in} 
\exists \vec{u}_L ((M_L,\vec{u}_L) \sat [\vec{Y} \gets \vec{y}](\V_L = \vec{v}_L))\})\\
  = &{\Pr}_L^{\vec{Y} \gets \vec{y}}(\{\vec{v}_L: \tau(\vec{v}_L) =
  \vec{v}_H\})\\
 = &\tau({\Pr}_L^{\vec{Y} \gets \vec{y}})(\vec{v}_H),
\end{array}
$$
}%end
$$\begin{array}{lll}
  &{\Pr}_H^{\omega(\do{\vec{Y} \gets \vec{y}})}(\vec{v}_H) \\
  = & {\Pr}_H(\{\vec{u}_H:
  M_H(\vec{u}_H,\omega(\vec{Y} \gets \vec{y})) =
  \vec{v}_H\})\\
  = &{\Pr}_L(\{\vec{u}_L: M_H(\tau_{\U}(\vec{u}_L),\omega(\vec{Y} \gets \vec{y})) =
  \vec{v}_H)\})\\
  = &{\Pr}_L(\{\vec{u}_L:  \tau(M_L(\vec{u}_L,\vec{Y} \gets \vec{y})) = \vec{v}_H\})\\
= &{\Pr}_L(\{\vec{v}_L: \tau(\vec{v}_L) =
%sander6: I've shortened it for space, but now the column is too
%wide. However, if we move everything to the left and decrease the
%space between the = and the formulas it should work, I just don't
%know how to do that. 
\vec{v}_H  \mbox{ and } %\\ &\hspace{.3in} 
%joe2: they will complain about the line being overlong, so I just
%broke the line.  WE have room
\\&\hspace{.35in}
\exists \vec{u}_L (M_L(\vec{u}_L,\vec{Y} \gets \vec{y})= \vec{v}_L)\})\\
  = &{\Pr}_L^{\vec{Y} \gets \vec{y}}(\{\vec{v}_L: \tau(\vec{v}_L) =
  \vec{v}_H\})\\
 = &\tau({\Pr}_L^{\vec{Y} \gets \vec{y}})(\vec{v}_H),
\end{array}
$$
as desired.
\eprf

{\bf Proof of Theorem~\ref{thm:abstuniform}:}
This follows immediately from Theorem~\ref{thm:uniformchar} once
we show that $\omega_\tau: \I(\V_L) \rightarrow \I(\V_H)$ is
order-preserving (it is surjective by definition).
Suppose that
$\omega_{\tau}(\vec{X} \gets \vec{x})= \vec{Y} \gets \vec{y}$ and
$\vec{X} \gets \vec{x} 
\preceq_{M_L}   \vec{X}' \gets \vec{x}'$.  Thus $\vec{X}$ is a subset of
$\vec{X}'$ and $\vec{x}$ is the corresponding subset of $\vec{x}'$.
Suppose that 
$\omega_{\tau}(\vec{X}' \gets \vec{x}')=\vec{Y}' \gets \vec{y}'$.  We
must show that $\vec{Y} 
\gets \vec{y}  \preceq_{M_H}  \vec{Y}' \gets \vec{y}'$.

By definition of $\Rest$,  $\Rest(\vec{x}') \subseteq \Rest(\vec{x})$. So
$\tau(\Rest(\vec{x}')) \subseteq \tau(\Rest(\vec{x}))$. But 
$\tau(\Rest(\vec{x}))=\Rest(\vec{y})$ and
$\tau(\Rest(\vec{x}')=\Rest(\vec{y}')$;  therefore 
$\Rest(\vec{y}') \subseteq \Rest(\vec{y})$. It immediately follows
that $\vec{Y} \gets \vec{y}  \preceq_{M_H}  \vec{Y}' \gets \vec{y}'$. 
\eprf

\bibliographystyle{aaai}
\bibliography{joe}

\end{document}